\documentclass[journal=jacsat,manuscript=article]{achemso}

\usepackage[version=3]{mhchem}
\usepackage{blindtext}
\usepackage{graphicx}

\title{Effect of structure-based training on 3D localization precision and quality}
\author{Armin Abdehkahka}
\author{Craig Snoeyink}
\email{craigsno@buffalo.edu}
\phone{+1 (716) 645-1468}
\affiliation[University at Buffalo]
{Department of Mechanical and Aerospace Engineering, University at Buffalo}
\begin{document}
\begin{abstract}
This study introduces a structural-based training approach for CNN-based algorithms in single-molecule localization microscopy (SMLM) and 3D object reconstruction. We compare this approach with the traditional random-based training method, utilizing the LUENN package as our AI pipeline. The quantitative evaluation demonstrates significant improvements in detection rate and localization precision with the structural-based training approach, particularly in varying signal-to-noise ratios (SNRs). Moreover, the method effectively removes checkerboard artifacts, ensuring more accurate 3D reconstructions. Our findings highlight the potential of the structural-based training approach to advance super-resolution microscopy and deepen our understanding of complex biological systems at the nanoscale.
\end{abstract}


\maketitle

\section{Introduction}
Single Molecule Localization Microscopy (SMLM) is a revolutionary technique in Super-resolution Microscopy that surpasses the diffraction limit, providing enhanced imaging resolution \citep{abbe1873beitrage}. This breakthrough allows researchers to study cellular functions relevant to both health and disease with unprecedented detail. Among various super-resolution imaging techniques, SMLM stands out by offering the highest achievable resolution, ranging from 20 to 30 nm, using relatively simple experimental equipment \citep{owen2013imaging}.

Despite its exceptional resolution, SMLM comes with challenges in terms of data analysis and imaging speed compared to other super-resolution techniques like STED microscopy \citep{hell1994breaking}, SIM \citep{gustafsson2000surpassing}, and NSOM \citep{vobornik2008fluorescence}. SMLM data analysis is complex and time-consuming, and the imaging speed is relatively slower, requiring several minutes to capture a complete dataset.

In SMLM, localization precision, and data interpretation heavily rely on the sparsity of fluorophores, ensuring well-separated point spread functions (PSFs). Achieving this requires time-separating frames and activating individual fluorophores. While increasing the fluorophore density per frame can enhance acquisition speed, it also introduces limitations. Higher fluorophore density can lead to PSF overlap, resulting in reduced detection accuracy and localization precision \citep{decode}. Additionally, high-density frames may produce artifacts like false structures, artificial sharpening, and checkerboard artifacts \citep{marsh2018artifact, decode}.

Given the demand for fast and accurate analysis methods that eliminate artifacts, especially in high-density frames, the development of an improved algorithm is highly sought after \citep{marsh2018artifact}. This advancement would enable precise localization in ultra-high densities, facilitating more accurate quantification and analysis of dynamic events, such as protein interactions in membrane fluidity analysis \citep{caetano2015miisr} within sub-cellular structures. Moreover, this progress in SMLM technology would have broad applications, including drug discovery, where a deeper understanding of protein interactions is essential for designing effective therapies.

Traditional mathematical-based localization algorithms, such as Maximum Likelihood Estimation (MLE) \citep{MLE} and non-linear least squares (LS) \citep{LS}, treat each point spread function (PSF) independently without considering their surroundings. While effective for sparse emitters with non-overlapping PSFs, these methods suffer from reduced precision when PSFs overlap, as they are unable to extract patterns from overlapping PSFs and leverage combined information for improved accuracy.

Recent advances in SMLM have seen significant progress in addressing these limitations by utilizing Convolutional Neural Networks (CNNs). State-of-the-art CNN-based algorithms, such as DeepLoco \citep{deeploco}, DeepSTORM3D \citep{deepstorm3d}, DECODE \citep{decode}, and LUENN \citep{luenn}, have achieved remarkable improvements in analysis times and localization accuracy. These deep-learning approaches excel at handling large numbers of emitters without a significant increase in computational time. However, highly overlapped patterns still pose challenges and can compromise reconstruction quality in scenarios with dense emitter distributions.

The performance of a CNN-based localization algorithm is heavily reliant on the training method used. While supervised learning is a common approach for robust training, it necessitates a large training dataset to avoid overfitting. However, in SMLM, obtaining real experimental frames with corresponding ground-truth data is limited, posing a challenge for training. To address this challenge, researchers often employ a reasonable frame generative model capable of reproducing data as closely as possible to real frames \citep{deeploco,deepstorm3d,decode}. The success of the AI model greatly depends on the accuracy of this generative model, and any mismatch between the simulated and experimental data could lead to reduced performance.

Frame simulation is a crucial three-step process in training a model for SMLM and 3D reconstruction of biological samples. Firstly, candidate seeds are randomly activated in a 3D grid domain, and their locations are selected within a confined domain both laterally (within the frame size) and axially (within the depth range). Next, the generative model employs the Point Spread Function (PSF) model to fill the frame with PSF distributions. Finally, camera noise is applied to simulate realistic frames that closely resemble experimental data.

While recent works on generative models have performed satisfactorily in PSF engineering modeling and camera noise estimation, less attention has been given to the sampling methods. Improving the sampling methods in frame simulation is an area of potential advancement to enhance the overall performance and accuracy of CNN-based localization algorithms in SMLM. In the context of seed sampling for frame simulation, the traditional approach adopted by researchers is the Complete Spatial Randomness (CSR) hypothesis. This method assumes that the points in the dataset are distributed randomly and independently throughout the study area, without any specific spatial patterns or interactions. Essentially, the points follow a homogeneous Poisson process, where the probability of finding a point at any location within the study area remains uniform and is not influenced by the presence or absence of other points.

The sampling method based on the CSR hypothesis is widely used as a baseline reference for evaluating deviations from randomness and uncovering any underlying spatial structures or dependencies present in real-world point datasets. By comparing the performance of other sampling methods to CSR, researchers can assess the effectiveness and efficiency of their frame simulation approaches in accurately representing and reproducing experimental data.

In the context of Single-Molecule Localization Microscopy (SMLM) and 3D reconstruction, the primary objective is to achieve rapid and accurate reconstruction of 3D objects. Given the specific goal of accurate 3D reconstruction, the CSR method, which assumes random and independent distributions, does not capture the intricacies and spatial relationships present in real 3D structures. Instead, specialized sampling methods that consider the actual 3D structure and interactions between points are necessary to produce realistic frame simulations. This is because SMLM datasets consist of collections of 3D points that are correlated in location due to the surface or geometry of an object or environment.

In our research, we propose a structure-based training approach to enhance the performance of CNN-based algorithms, especially in ultra-high emitter densities. Structure-based training involves considering the underlying structure and relationships between neighboring PSFs, instead of treating data as independent samples. By leveraging the contextual information and correlations between data points, the algorithm can make more informed predictions and capture dependencies present in the structured data. Sampling from cloud points, where overlapping patterns have meaningful relationships, allows the CNN to reconstruct a more accurate 3D view of the structure.

Incorporating structure-based training can provide several advantages, including improved accuracy, robustness, and generalization performance of the trained model. By advancing the localization precision of CNN-based algorithms in SMLM, we can unlock the full potential of SMLM and enhance our understanding of molecular interactions in biological systems. This research has the potential to significantly impact the field of super-resolution microscopy and contribute to breakthroughs in various biological and medical research areas.

\section{Methods}
\begin{figure}[!h]
\centering
\includegraphics[width=.9 \textwidth]{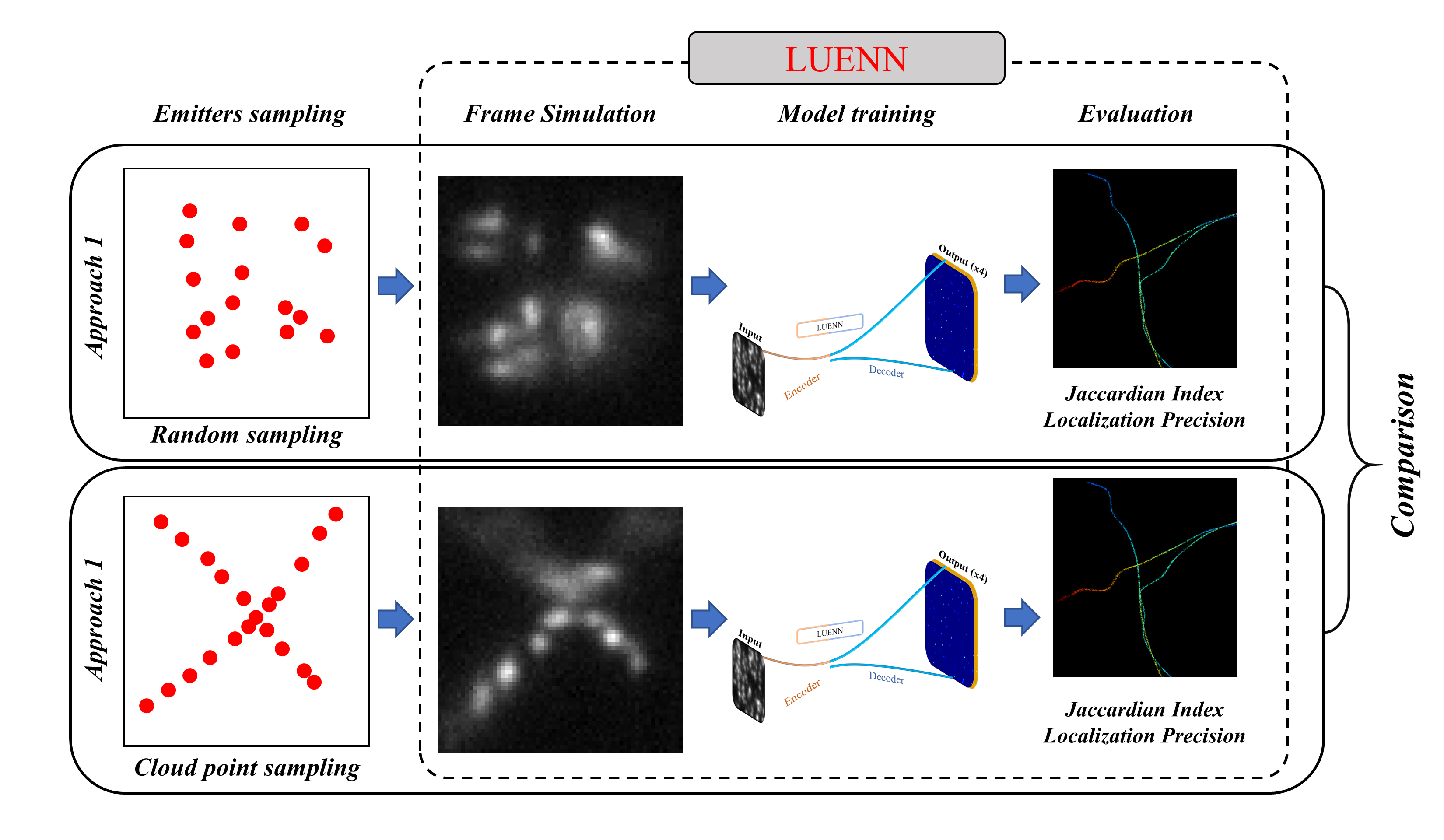}
\caption{\textbf{Comparison Workflow}. Contrasting two training methodologies rooted in random emitter sampling (Approach 1) and structure-based sampling (Approach 2). In both approaches, the LUENN pipeline is employed for frame simulation, incorporating Astigmatism modality, model training, and evaluation, with calculations performed for Jaccardian Index and precision assessments.}\label{fig:Figure1}
\end{figure}

In our study, we aimed to compare the effects of two different seed sampling methods on the performance of a CNN-based localization algorithm. To conduct the experiments, we utilized the LUENN package, which served as our AI pipeline, guiding us through each step of the process, including emitters sampling, frame simulation, model selection, model training, and performance evaluation. The overall process is illustrated in Figure\ref{fig:Figure1}.

We utilized the Astigmatism modality in both approaches and trained the model on three distinct background-to-noise ratios: high, medium, and low. These three refer to mean photon counts of 1000, 5000, and 20,000 with background levels of 10, 50, and 200 photons per pixel, respectively. The frame generative model used in LUENN was based on the method presented by \cite{decode}, and the training procedure was comprehensively explained in our previous work \citep{luenn}.

The primary focus of our investigation was to compare two different approaches in the emitters sampling step. In the first approach, we trained LUENN by sampling emitters based on the Complete Spatial Randomness (CSR) hypothesis. In the second approach, emitters were sampled from randomly generated point clouds, which aim to represent real 3D structures.

After completing the full training of both models, we proceeded to compare their reconstruction and localization performance using artificially generated framesets of Microtubules, utilizing the ground-truth data from the MT0 datasets provided by the SMLM Challenge 2016. The ground-truth emitters data can be accessed through the following LINK. However, it's worth noting that the challenge only provided framesets for two specific frame densities, limiting our comparison's scope.

To overcome this limitation, we generated an additional nine sets of frames with varying nominal densities, ranging from 0.38 to 13.0 emitters per frame. It's important to mention that these densities are not the actual density that is formulated as the number of emitters per unit area in the frame. In the frames where the points' distribution follows a specific structure, such as a line or a surface, the actual density may not accurately reflect the difficulty of localization. 

We employed Ripley's method \citep{ripley} to calculate the average minimum distance between the emitters on the simulated frames to address this. Using cross-correlation, we then identified the frame density with a similar average minimum distance, in which the seeds were uniformly distributed. This corresponding frame density was considered the ``nominal density'' of the frame, providing a more accurate and broadly applicable quantification metric.

Both models followed the same on-the-fly training procedure, where training frames were generated randomly during the training process. For training LUENN with the CSR method, the seeds were randomly sampled across the entire 3D domain without any constraints. On the other hand, for the structure-based training, we created three-helix microtubules that spanned the 3D domain, each containing 5000 seeds. From these sampled point clouds, we selected candidate emitters, considering the challenges posed by overlapping PSFs along a spiral curve. It's important to note that the sampled microtubules were only used for one frame of the training data and the microtubule structures were continuously changed throughout the training process.

\begin{figure}[!h]
\centering
\includegraphics[width=.9 \textwidth]{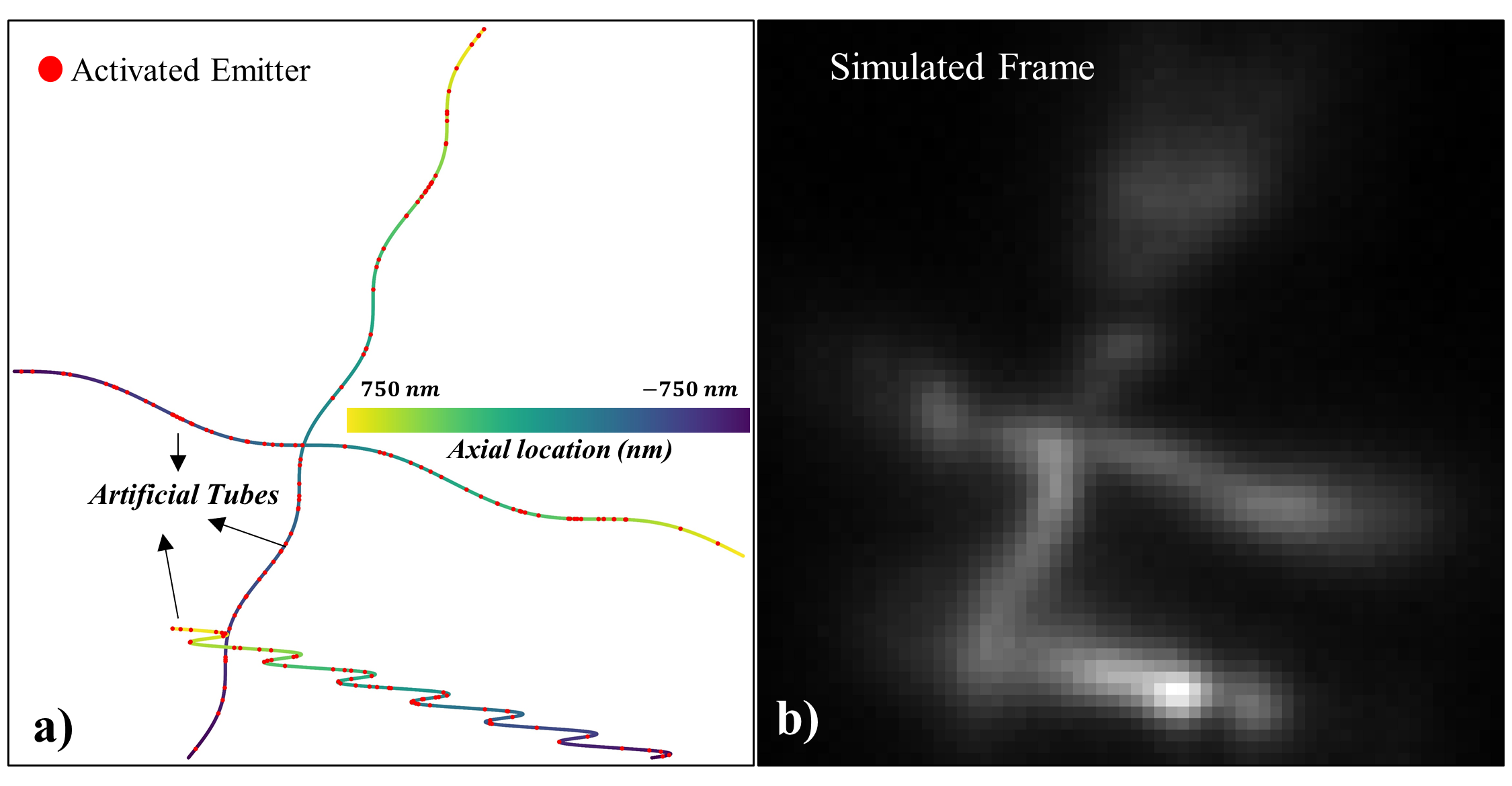}
\caption{Investigating two sampling approaches on the performance of the LUENN on 3D reconstruction. Approach 1: sampling emitters randomly without prior information of the real object. Approach 2: Cloud point sampling to give the model rationale behind the overlapping PSFs. both approaches share the same method for frame simulation, model training, and performance evaluation that all are part of the LUENN pipeline.}\label{fig:Figure2}
\end{figure}

In Figure \ref{fig:Figure2}, we provide two examples of training frames along with their corresponding labeled frames. These examples showcase the effectiveness of the structure-based training approach in addressing the challenges of emitter localization, particularly in cases with overlapping PSFs, thus contributing to the improved performance of the CNN-based localization algorithm.

\section{Results and discussion}
\begin{figure}[!h]
\centering
\includegraphics[width=.9 \textwidth]{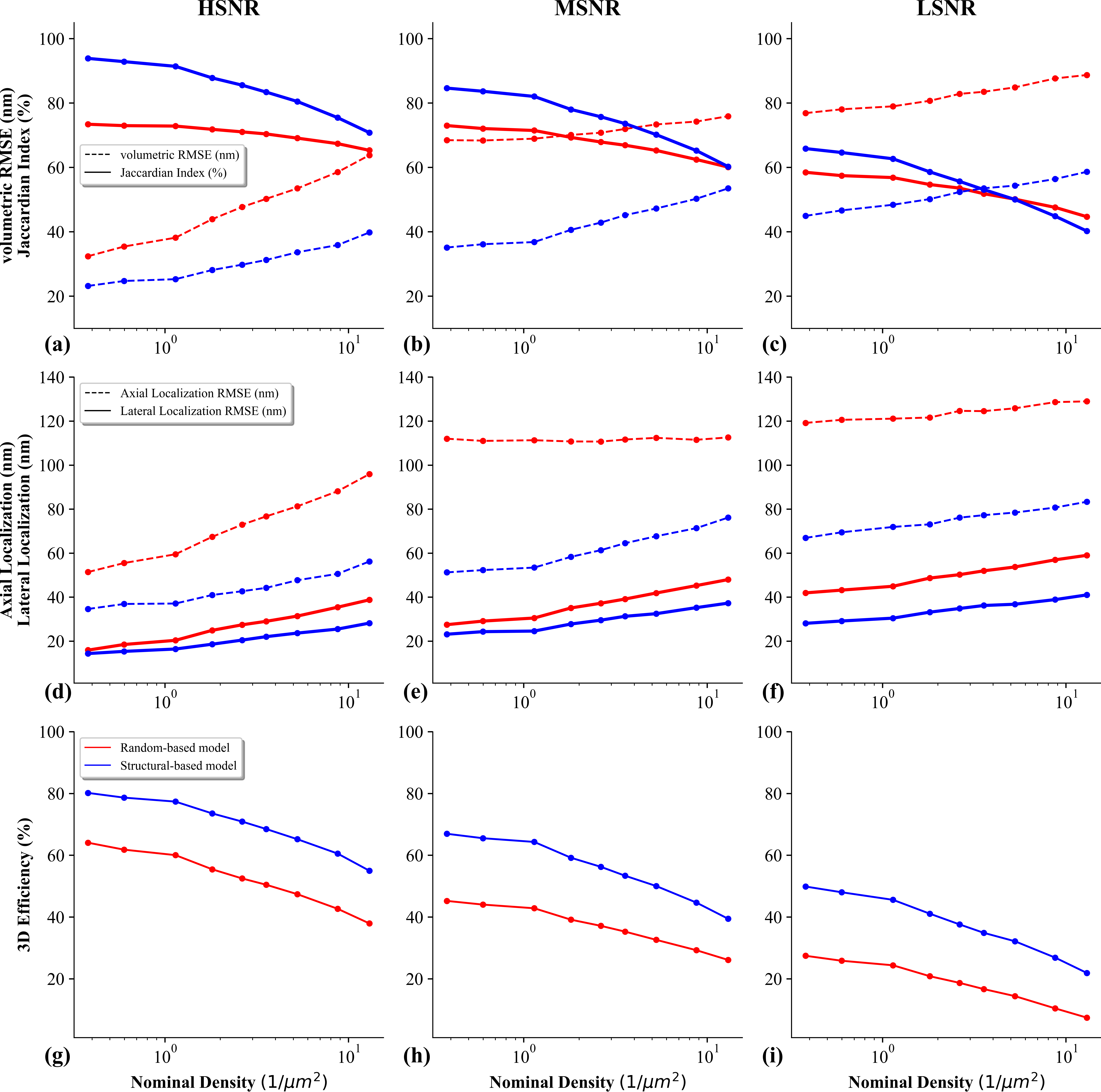}
\caption{
Quantitative comparison of training methods, encompassing both random and structural-based approaches, across varying signal-to-noise ratios (SNRs) – high, medium, and low. \textbf{a-c}. Depicting the Jaccardian index and volumetric RMSE as functions of frame densities. \textbf{e-f}. Illustrating the lateral and axial RMSE trends relative to frame densities. \textbf{g-i}. Presenting the 3D efficiency trends in relation to frame densities.}\label{fig:Figure3}
\end{figure}

For the quantitative evaluation, we employed two metrics, namely the Jaccardian index (JI) and the root-mean-square error (RMSE), to assess the detection rate and localization accuracy in the X, Y, and Z directions in reconstructing the MT0 dataset. Figure \ref{fig:Figure3}a to c presents the Jaccardian index and volumetric RMSE plotted against the frame densities. The results demonstrate a significant improvement in both detection rate and localization precision when using the structural-based training approach. In high signal-to-noise ratio (SNR) conditions, the structural-based model exhibited an average of 20\% improvement in the detection rate and 35\% improvement in RMSE compared to the traditional random-based training model. 

Similar improvements are observed in medium and low SNR scenarios, as shown in Figures \ref{fig:Figure3}b and \ref{fig:Figure3}c. In these cases, the structural-based trained model outperformed the random-based trained model with 7.2\% and 2.2\% higher detection rates, respectively. Additionally, on average, the volumetric RMSE was 28.3 nm and 30.7 nm lower in the structural-based training approach for medium and low SNRs, respectively. These results demonstrate the robustness and superior performance of the structural-based training approach in various SNR conditions, making it a promising method for enhancing the accuracy and precision of single-molecule localization microscopy (SMLM) in 3D object reconstruction.

Figures \ref{fig:Figure3}d to f present the lateral and axial localization RMSE for high, medium, and low SNRs, respectively. These results underscore the substantial improvement in localization precision, particularly in the depth (Z) direction, further validating the robustness of the structural-based training approach in achieving accurate 3D object reconstruction, which is the primary goal of SMLM. An interesting trend in the results is that as the nominal density increases, our new training method exhibits even better results, as indicated by the increasing difference between the plots for both lateral and axial localization RMSE. This finding highlights the effectiveness of the structural-based training approach in handling varying emitter densities and its potential to provide superior localization accuracy across a wide range of imaging conditions.

Figures \ref{fig:Figure3}g to i illustrate the 3D efficiency of the two methods for high, medium, and low SNRs, respectively. The results clearly demonstrate a significant improvement in 3D efficiency by adopting the structural-based training approach. On average, the performance of LUENN improved by 20\% across all three levels of SNRs. This enhancement in 3D efficiency showcases the superiority of the structural-based training method in achieving more accurate and reliable 3D reconstructions in single-molecule localization microscopy.

\begin{figure}[!h]
\centering
\includegraphics[width=.9 \textwidth]{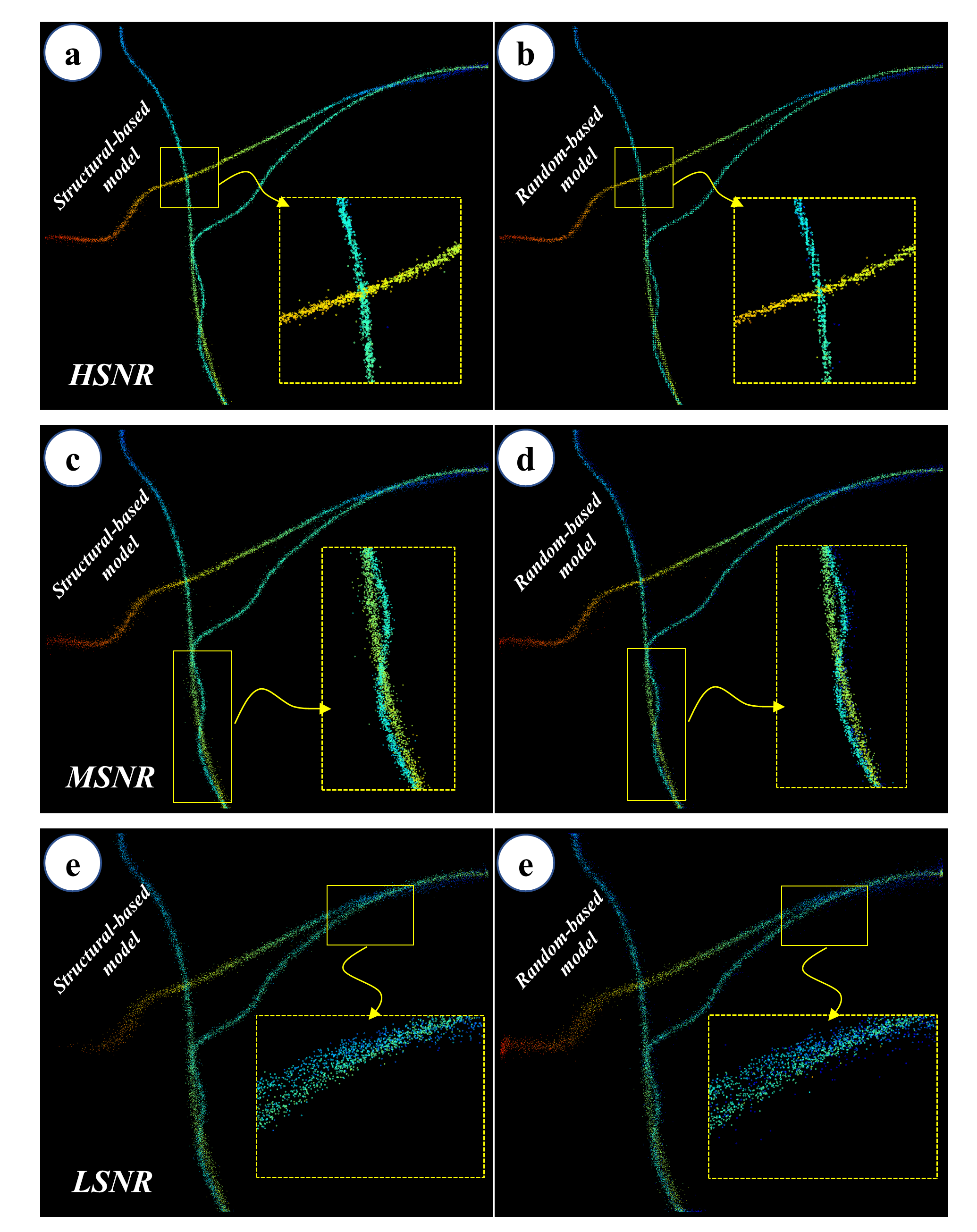}
\caption{Visual evaluation of reconstruction quality under low-density frame conditions (with a nominal density of 0.38 emitters per frame) across diverse signal-to-noise ratios (SNRs), encompassing high, medium, and low levels. Panels \textbf{a-b} illustrate the reconstructions under high SNR, \textbf{c-d} depict the reconstructions under medium SNR, and \textbf{e-f} showcase the reconstructions under low SNR.}\label{fig:Figure4}
\end{figure}

To visually assess the quality of the reconstructed Microtubules, we have provided reconstruction of low density frames, nominal density equal to 0.38 emitters per frame, at high, medium, and low SNRS in Figures \ref{fig:Figure4} a to f. Notably, the structural-based training approach has demonstrated its effectiveness in removing checkerboard artifacts, which are pixel-level biases resulting in grid-like patterns in the reconstructed structures. These artifacts can significantly impact the accuracy of 3D reconstructions and may lead to misleading interpretations of biological structures.

By successfully mitigating these checkerboard artifacts, the structural-based training approach ensures a more reliable and accurate reconstruction of 3D structures, thereby advancing the capabilities of SMLM in studying complex biological systems. This improvement in reconstruction quality has important implications for understanding the spatial organization of cellular components and interactions within subcellular structures, contributing to further breakthroughs in the field of super-resolution microscopy and its applications in biological and medical research.

Finally, structure-based training does not negatively impact the localization of isolated emitters. As Figure \ref{fig:Figure3} shows, even at low emitter densities the structure-based training method demonstrates significant improvement in both localization accuracy and precision. This means that even in the absence of multiple emitters in an image, i.e. when the neural network can't utilize additional information about structure, the neural network performs better at localization. 

\section{Conclusion}
In conclusion, our study has demonstrated the efficacy of the structural-based training approach in significantly improving the performance of CNN-based algorithms for single-molecule localization microscopy (SMLM) and 3D object reconstruction. By utilizing two key metrics, the Jaccardian index as a measure of localization accuracy (JI) and root-mean-square error (RMSE) as a measure of precision, we have quantitatively shown the superiority of the structural-based model over the traditional random-based training model.

The structural-based training approach exhibited remarkable improvements in both detection rate and localization precision across a wide range of signal-to-noise ratios (SNRs), outperforming the random-based trained model in high, medium, and low SNR conditions. With an average of 20\% higher detection rate and 35\% lower RMSE in high SNR, and 7.2\% and 5.6\% higher detection rates in medium and low SNRs, respectively, the structural-based model showcases its robustness and versatility. These advantages exist even at very low emitter densities where the structure is not obvious in every image. 

Notably, the structural-based training approach excelled in achieving superior localization precision, especially in the depth (Z) direction, which is crucial for accurate 3D object reconstruction. The increasing difference between lateral and axial localization RMSE plots with higher nominal density further highlights the method's effectiveness in handling varying emitter densities and enhancing localization accuracy.

The visual assessment of reconstructed Microtubules demonstrated that the structural-based training approach effectively eliminated checkerboard artifacts, a common issue affecting the accuracy of 3D reconstructions. This improvement ensures more reliable and accurate representations of biological structures, enhancing the capabilities of SMLM in studying complex subcellular systems.

Overall, our findings suggest that the structural-based training approach holds great promise for advancing the field of super-resolution microscopy and its applications in biological and medical research. By improving localization precision and reconstruction quality, this novel approach can lead to deeper insights into the spatial organization of cellular components and interactions, ultimately contributing to significant breakthroughs in understanding complex biological processes. As a result, the structural-based training approach can have a transformative impact on the study of biological systems at the nanoscale, offering new avenues for exploration and discoveries.

\begin{acknowledgement}
We acknowledge the Center for Computational Research at the University at Buffalo for providing the computational support needed for training the models and for supplying the results of this study.
\end{acknowledgement}
\bibliography{main}
\end{document}